# Heuristic Search Value Iteration for POMDPs

**Trey Smith** and **Reid Simmons**
Robotics Institute, Carnegie Mellon University
{trey,reids}@ri.cmu.edu

## Abstract

We present a novel POMDP planning algorithm called heuristic search value iteration (HSVI). HSVI is an anytime algorithm that returns a policy and a provable bound on its regret with respect to the optimal policy. HSVI gets its power by combining two well-known techniques: attention-focusing search heuristics and piecewise linear convex representations of the value function. HSVI's soundness and convergence have been proven. On some benchmark problems from the literature, HSVI displays speedups of greater than 100 with respect to other state-of-the-art POMDP value iteration algorithms. We also apply HSVI to a new rover exploration problem 10 times larger than most POMDP problems in the literature.

## 1 INTRODUCTION

Partially observable Markov decision processes (POMDPs) constitute a powerful probabilistic model for planning problems that include hidden state and uncertainty in action effects. There are a wide variety of solution approaches. To date, problems of a few hundred states are at the limits of tractability.

The present work gathers a number of threads in the POMDP literature. Our HSVI algorithm draws on prior approaches that combine heuristic search and value iteration [Washington, 1997, Geffner and Bonet, 1998], and a multitude of algorithms that employ a piecewise linear convex value function representation and gradient backups [Cassandra et al., 1997, Pineau et al., 2003]. It keeps compact representations of both upper and lower bounds on the value function [Hauskrecht, 1997]. Making use of these bounds, HSVI incorporates a novel *excess uncertainty* observation heuristic that empirically outperforms the usual sampling, and allows us to derive a theoretical bound on time complexity.

By employing all of these techniques, HSVI gains a number of benefits. Its use of heuristic search forward from an initial belief (aided by the new observation heuristic) avoids unreachable or otherwise irrelevant parts of the belief space. Its representations for both bounds are compact and well-suited to generalizing local updates: improving the bounds at a specific belief also improves them at neighboring beliefs.

Some weaknesses of HSVI are that it is relatively complicated, and its upper bound updates are a source of major overhead that only becomes worthwhile on the larger problems we studied.

This paper describes HSVI, discusses its soundness and convergence, and compares its performance with other state-of-the-art value iteration algorithms on four benchmark problems from the literature. On some of these problems, HSVI displays speedups of greater than 100. We provide additional results on the new *RockSample* rover exploration problem. Our largest instance of *RockSample* has 12,545 states, 10 times larger than most problems in the scalable POMDP literature.

## 2 POMDP INTRODUCTION

A POMDP models an agent acting under uncertainty. At each time step, the agent selects an action that has some stochastic result, then receives a noisy observation. The sequence of events can be viewed as a tree structure (fig. 1). Nodes of the tree are points where the agent must make a decision. We label each node with the belief $b$ that the agent would have if it reached that node. The root node is labeled with the initial belief, $b_0$. Starting from node $b$, selecting action $a$, and receiving observation $o$, the agent proceeds to a new belief $\tau(b, a, o)$, corresponding to one of the children of $b$ in the tree structure.

Formally, the POMDP is a tuple $\langle \mathcal{S}, \mathcal{A}, \mathcal{O}, T, O, R, \gamma, b_0 \rangle$, where $\mathcal{S}$ is the set of states, $\mathcal{A}$ the set of actions, $\mathcal{O}$ is the set of observations, $T$ is the stochastic transition function, $O$ is the stochastic observation function, $R$ is the reward function, $\gamma < 1$ is the discount factor, and $b_0$ is the agent's belief about the initial state. Let $s_t$, $a_t$, and $o_t$ denote, respectively, the state, action, and observation at time $t$. Then we define

$$\begin{aligned} b_0(s) &= \Pr(s_0 = s) \\ T(s, a, s') &= \Pr(s_{t+1} = s' \mid s_t = s, a_t = a) \\ O(s, a, o) &= \Pr(o_t = o \mid s_{t+1} = s, a_t = a). \end{aligned}$$

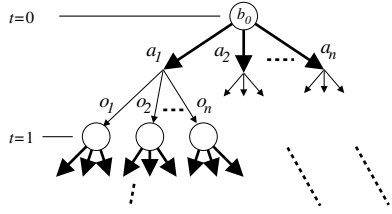

Figure 1: POMDP tree structure.

Let $\mathbf{a}^t = \{a_0, \ldots, a_t\}$ denote the history of actions up to time $t$, and similarly define $\mathbf{o}^t$. At time $t+1$, the agent does not know $s_{t+1}$, but does know the initial belief $b_0$, and the history $\mathbf{a}^t$ and $\mathbf{o}^t$. The agent can act optimally on this information by conditioning its policy on its current belief at every step. The belief is recursively updated as follows:

$$b_{t+1} = \tau(b_t, a_t, o_t),$$

If $b' = \tau(b, a, o)$, then

$$b'(s') = \eta O(s', a, o) \sum_s T(s, a, s') b(s),$$

where $\eta$ is a normalizing constant. The agent's policy $\pi$ specifies an action $\pi(b)$ to follow given any current belief $b$.

The (expected) long-term reward for a policy $\pi$, starting from a belief $b$, is defined to be

$$J^\pi(b) = E\left[\sum_{t=0}^\infty \gamma^t R(s_t, a_t) \mid b, \pi\right].$$

The optimal POMDP planning problem is to compute a policy $\pi^*$ that optimizes long-term reward.

$$\pi^* = \operatorname*{argmax}_\pi J^\pi(b_0)$$

The usual goal for approximate POMDP planning is to minimize the *regret* of the returned policy $\pi$, defined to be

$$\operatorname{regret}(\pi, b) = J^{\pi^*}(b) - J^\pi(b)$$

In particular, we focus on minimizing $\operatorname{regret}(\pi, b_0)$ for the initial belief $b_0$ specified as part of the problem.

## 3 HEURISTIC SEARCH VALUE ITERATION

HSVI is an approximate POMDP solution algorithm that combines techniques for heuristic search with piecewise linear convex value function representations. HSVI stores upper and lower bounds on the optimal value function $V^*$. Its fundamental operation is to make a local update at a specific belief, where the beliefs to update are chosen by exploring forward in the search tree according to heuristics that select actions and observations.

HSVI makes asynchronous (Gauss-Seidel) updates to the value function bounds, and always bases its heuristics on the most recent bounds when choosing which successor to visit. It uses a depth-first exploration strategy. Beyond the usual memory vs. time trade-off, this choice makes sense because a breadth-first heuristic search typically employs a priority queue, and propagating the effects of asynchronous bounds updates to the priorities of queue elements would create substantial extra overhead.

We refer to the lower and upper bound functions as $\underline{V}$ and $\bar{V}$, respectively. We use the interval function $\hat{V}$ to refer to them collectively, such that

$$\begin{aligned} \hat{V}(b) &= [\underline{V}(b), \bar{V}(b)] \\ \operatorname{width}(\hat{V}(b)) &= \bar{V}(b) - \underline{V}(b) \end{aligned}$$

HSVI is outlined in algs. 1 and 2. The following subsections describe aspects of the algorithm in more detail.

### 3.1 VALUE FUNCTION REPRESENTATION

Most value iteration algorithms focus on storing and updating the lower bound. The *vector set* representation is commonly used. The value at a point $b$ is the maximum projection of $b$ onto a finite set $\Gamma_{\underline{V}}$ of vectors $\alpha$:

$$\underline{V}(b) = \max_{\alpha \in \Gamma_{\underline{V}}} (\alpha \cdot b).$$

For finite-horizon POMDPs, a finite vector set can represent $V^*$ exactly [Sondik, 1971]. Even for the discounted infinite-horizon formulation, a finite vector set can approximate $V^*$ arbitrarily closely. Equally important, when the value function is a lower bound, it is easy to perform a local update on the vector set by adding a new vector.

Unfortunately, if we represent the upper bound with a vector set, updating by adding a vector does not have the desired effect of improving the bound in the neighborhood of the local update. To accommodate the need for updates, we use a *point set* representation for the upper bound. The value at a point $b$ is the projection of $b$ onto the convex hull formed by a finite set $\Upsilon_{\bar{V}}$ of belief/value points $(b_i, \bar{v}_i)$. Updates are performed by adding a new point to the set.

The projection onto the convex hull is calculated with a linear program (LP). This upper bound representation and

**Algorithm 1** $\pi = \text{HSVI}(\epsilon)$.

HSVI($\epsilon$) returns a policy $\pi$ such that $\text{regret}(\pi, b_0) \leq \epsilon$.[a]

1. Initialize the bounds $\hat{V}$.
2. While $\text{width}(\hat{V}(b_0)) > \epsilon$, repeatedly invoke $\text{explore}(b_0, \epsilon, 0)$.
3. Having achieved the desired precision, return the direct-control policy $\pi$ corresponding to the lower bound.

[a]In fact, $\pi$ can be executed starting at any belief $b$. In general, $\text{regret}(\pi, b) \leq \text{width}(\hat{V}(b))$, which is guaranteed to be less than $\epsilon$ only at $b_0$.

**Algorithm 2** $\text{explore}(b, \epsilon, t)$.

explore recursively follows a single path down the search tree until satisfying a termination condition based on the width of the bounds interval. It then performs a series of updates on its way back up to the initial belief.

1. If $\text{width}(\hat{V}(b)) \leq \epsilon \gamma^{-t}$, return.
2. Select an action $a^*$ and observation $o^*$ according to the forward exploration heuristics.
3. Call $\text{explore}(\tau(b, a^*, o^*), \epsilon, t+1)$.
4. Locally update the bounds $\hat{V}$ at belief $b$.

LP technique was suggested in [Hauskrecht, 2000], but in that work LP projection seems to have been rejected without testing on time complexity grounds. Note that with the high dimensionality of the belief space in our larger problems, LP projection is far more efficient than explicitly calculating the convex hull: an explicit representation would not even fit into available memory. We solve the LP using the commercial ILOG CPLEX software package.

### 3.2 INITIALIZATION

HSVI requires initial bounds, which we would like to have the following properties:

1. *Validity:* $\underline{V}_0 \leq V^* \leq \bar{V}_0$.[1]
2. *Uniform improvability:* This property is explained in the section on theoretical results.
3. *Precision:* The bounds should be fairly close to $V^*$.
4. *Efficiency:* Initialization should take a negligible proportion of the overall running time.

The following initialization procedure meets these requirements. We calculate $\underline{V}_0$ using the blind policy method

[1]Throughout this paper, inequalities between functions are universally quantified, i.e., $V \leq V'$ means $V(b) \leq V'(b)$ for all $b$.

[Hauskrecht, 1997]. Let $\pi_a$ be the policy of always selecting action $a$. We can calculate a lower bound $\underline{R}_a$ on the long-term reward of $\pi_a$ by assuming that we are always in the worst state to choose action $a$ from.

$$\underline{R}_a = \sum_{t=0}^{\infty} \gamma^t \min_s R(s, a) = \frac{\min_s R(s, a)}{1 - \gamma}$$

We select the tightest of these bounds by maximizing.

$$\underline{R} = \max_a \underline{R}_a$$

Then the vector set for the initial lower bound $\underline{V}_0$ contains a single vector $\alpha$ such that every $\alpha(s) = \underline{R}$.

To initialize the upper bound, we assume full observability and solve the MDP version of the problem [Astrom, 1965]. This provides upper bound values at the corners of the belief simplex, which form the initial point set. We call the resulting upper bound $V_{\text{MDP}}$.

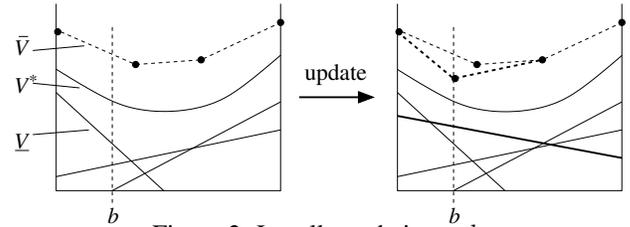

Figure 2: Locally updating at $b$.

### 3.3 LOCAL UPDATES

The Bellman update, $H$, is the fundamental operation of value iteration. It is defined as follows:

$$Q^V(b, a) = \sum_s R(s, a)b(s) + \gamma \sum_o \Pr(o \mid b, a) V(\tau(b, a, o))$$

$$HV(b) = \max_a Q^V(b, a)$$

$Q^V(b, a)$ can be interpreted as the value of taking action $a$ from belief $b$.

Exact value iteration calculates this update exactly over the entire belief space. HSVI, however, uses local update operators based on $H$. Because the lower and upper bound are represented differently, we have distinct local update operators $L_b$ and $U_b$. *Locally updating* at $b$ means applying both operators. To update the lower bound vector set, we add a vector. To update the upper bound point set, we add a point. The operators are defined as:

$$\Gamma_{L_b \underline{V}} = \Gamma_{\underline{V}} \cup \text{backup}(\underline{V}, b)$$
$$\Upsilon_{U_b \bar{V}} = \Upsilon_{\bar{V}} \cup (b, H\bar{V}(b)),$$

where $\text{backup}(\underline{V}, b)$ is the usual gradient backup, described in alg. 3.

**Algorithm 3** $\beta = \text{backup}(\underline{V}, b)$.

The backup function can be viewed as a generalization of the Bellman update that makes use of gradient information. The assignments are universally quantified, e.g., $\beta_{a,o}$ is computed for every $a$, $o$.

1. $\beta_{a,o} \leftarrow \text{argmax}_{\alpha \in \Gamma_V}(\alpha \cdot \tau(b, a, o))$
2. $\beta_a(s) \leftarrow R(s, a) + \gamma \sum_{o, s'} \beta_{a,o}(s') O(s', a, o) T(s, a, s')$
3. $\beta \leftarrow \text{argmax}_{\beta_a}(\beta_a \cdot b)$.

Fig. 2 represents the structure of the bounds representations and the process of locally updating at $b$. In the left side of the figure, the points and dotted lines represent $\bar{V}$ (upper bound points and convex hull). Several solid lines represent the vectors of $\Gamma_{\underline{V}}$. In the right side of the figure, we see the result of updating both bounds at $b$, which involves adding a new point to $\Upsilon_{\bar{V}}$ and a new vector to $\Gamma_{\underline{V}}$, bringing both bounds closer to $V^*$.

HSVI periodically prunes dominated elements in both the lower bound vector set and the upper bound point set. Pruning occurs each time the size of the relevant set has increased by 10% since the last pruning episode. This pruning frequency was not carefully optimized, but there is not much to be gained by tuning it, since we do note see substantial overhead either from keeping around up to 10% too many elements or from the pruning operation itself. For the lower bound, we prune only vectors that are *pointwise* dominated (i.e., dominated by a single other vector). This type of pruning does not eliminate all redundant vectors, but it is simple and fast. For the upper bound, we prune all dominated points, defined as $(b_i, \bar{v}_i)$ such that $H\bar{V}(b_i) < \bar{v}_i$.

### 3.4 FORWARD EXPLORATION HEURISTICS

This section discusses the heuristics that are used to decide which child of the current node to visit as the algorithm works its way forward from the initial belief. Starting from parent node $b$, HSVI must choose an action $a^*$ and an observation $o^*$: the child node to visit is $\tau(b, a^*, o^*)$.

Define the *uncertainty* at $b$ to mean the width of the bounds interval. Recalling that the regret of a policy returned by HSVI is bounded by the uncertainty at the root node $b_0$, our goal in designing the heuristics is to ensure that updates at the chosen child tend to reduce the uncertainty at the root.

First we consider the choice of action. Define the interval function $\hat{Q}$ as follows:

$$\hat{Q}(b, a) = [Q^{\underline{V}}(b, a), Q^{\bar{V}}(b, a)]$$

Fig. 3 shows the relationship between the bounds $\hat{Q}(b, a)$ on each potential action and the bounds $H\hat{V}(b)$ at $b$ after a

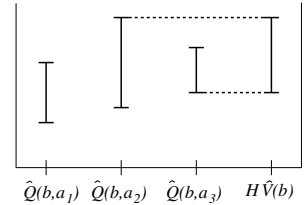

Figure 3: Relationship between $\hat{Q}(b, a_i)$ and $H\hat{V}(b)$.

Bellman update. We see that the $H\hat{V}(b)$ interval is determined by only two of the $\hat{Q}(b, a)$ intervals: the ones with the maximal upper and lower bounds. This relationship immediately suggests that, among the $\hat{Q}$ intervals, we should choose to update one of these two most promising actions. But which one? It turns out we can guarantee convergence only by choosing the action with the greatest *upper* bound.

$$a^* = \underset{a}{\text{argmax}}\, Q^{\bar{V}}(b, a).$$

This is sometimes called the IE-MAX heuristic [Kaelbling, 1993]. It works because, if we repeatedly choose an $a^*$ that is sub-optimal, we will eventually discover its sub-optimality when the $a^*$ upper bound drops below the upper bound of another action. However, if we were to choose $a^*$ according to the highest lower bound, we might never discover its sub-optimality, because further work could only cause its lower bound to rise.

Next we need to select an observation $o^*$. Consider the relationship between $\hat{Q}(b, a^*)$ and the bounds at the various child nodes $\tau(b, a^*, o)$ that correspond to different observations. From the Bellman equation, we have

$$\text{width}(\hat{Q}(b, a^*)) = \gamma \sum_o \Pr(o|b, a^*) \text{width}(\hat{V}(\tau(b, a^*, o))).$$

Note that this explains the termination criterion of explore, $\text{width}(\hat{V}(b)) \le \epsilon \gamma^{-t}$. Because the uncertainty at a node $b$ after an update is at most $\gamma$ times a weighted average of its child nodes, we have successively looser requirements on uncertainty at deeper nodes: we rely on the $\gamma$ factor at each layer on the way back up to make up the difference. Given these facts, we can define *excess uncertainty*

$$\text{excess}(b, t) = \text{width}(\hat{V}(b)) - \epsilon \gamma^{-t}$$

such that a node with negative excess uncertainty satisfies the explore termination condition. We say such a node is *finished*. Conveniently, the excess uncertainty at $b$ is at most a probability-weighted sum of the excess uncertainties at its children

$$\text{excess}(b, t) \le \sum_o \Pr(o|b, a^*) \text{excess}(\tau(b, a^*, o), t+1).$$

Thus we can focus on ensuring early termination by selecting the depth $t + 1$ child that most contributes to excess uncertainty at $b$:

$$o^* = \underset{o}{\text{argmax}} \left[ \Pr(o|b, a^*) \text{excess}(\tau(b, a^*, o), t+1) \right].$$

**Algorithm 4** $\pi = \text{AnytimeHSVI}()$.

AnytimeHSVI() is an anytime variant of HSVI. When interrupted, it returns a policy whose regret is bounded by the current value of $\text{width}(\hat{V}(b_0))$.

Implementation: As HSVI, but in step (2), in the call to $\text{explore}(b_0, \epsilon, 0)$, replace $\epsilon$ with $\zeta \, \text{width}(\hat{V}(b_0))$, where $\zeta < 1$ is a scalar parameter. Empirically, performance is not very sensitive to $\zeta$; we used $\zeta = 0.95$ in the experiments, which gives good performance.

Past heuristic search approaches have usually either sampled from $\Pr(o|b, a^*)$ or maximized weighted uncertainty rather than weighted *excess* uncertainty. We find the excess uncertainty heuristic to be empirically superior. In addition, this heuristic allows us to derive a time bound on HSVI convergence.

### 3.5 ANYTIME USAGE

The definition of HSVI($\epsilon$) given above assumes that we know in advance that we want a policy with regret bounded by $\epsilon$. In practice, however, we often do not know what a reasonable $\epsilon$ is for a given problem—we just want the algorithm to do the best it can in the available time. In support of this goal, we define a variant algorithm called AnytimeHSVI (alg. 4). Where HSVI uses a fixed $\epsilon$, AnytimeHSVI adjusts $\epsilon$ at each top-level call to explore, setting it to be slightly smaller than the current uncertainty at $b_0$. Instead of having a fixed finish line, we have a finish line that is always just ahead, receding as we approach.

AnytimeHSVI is used for all of the experiments in this paper. However, our theoretical analysis focuses on HSVI($\epsilon$), which is easier to handle mathematically.

## 4 THEORETICAL RESULTS

This section discusses some of the key soundness and convergence properties of HSVI($\epsilon$). The actual proofs are presented in [Smith and Simmons, 2004].

- The initial lower and upper bound value functions are *uniformly improvable*, meaning that applying $H$ brings them everywhere closer to $V^*$.
- If $\underline{V}$ is uniformly improvable, then the corresponding direct control policy $P_{\underline{V}}$ supports $\underline{V}$, meaning that $\underline{V} \leq J^{P_{\underline{V}}}$.[2]
- If $\bar{V}$ is uniformly improvable, then it is valid, in the sense that $V^* \leq \bar{V}$.

---

[2]Direct and lookahead control policies corresponding to a value function are discussed in, e.g., [Hauskrecht, 2000].

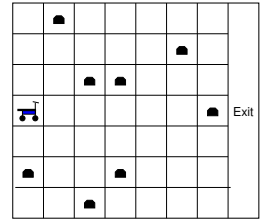

Figure 4: *RockSample*[7, 8].

- Our local update operators preserve uniform improvability. Thus, throughout the execution of HSVI, the current best policy $P_{\underline{V}}$ supports $\underline{V}$, and $\bar{V}$ is valid.
- Together, these facts imply that HSVI has valid bounds on the direct control policy, in the sense that $\underline{V} \leq J^{\pi_{\underline{V}}} \leq \bar{V}$. This validity holds throughout execution and everywhere in the belief space.
- The $\text{regret}(\pi, b_0)$ of the policy $\pi$ returned by HSVI($\epsilon$) is at most $\epsilon$. When AnytimeHSVI is interrupted, the $\text{regret}(\pi, b_0)$ of the current best policy $\pi$ is at most $\text{width}(\hat{V}(b_0))$.
- There is a finite depth
$$t_{\max} = \lceil \log_\gamma(\epsilon / \|\bar{V}_0 - \underline{V}_0\|_\infty) \rceil$$
such that all nodes with depth $t \geq t_{\max}$ are finished.
- The uncertainty at a node never increases (thus finished nodes never become unfinished).
- After each top-level call to explore, at least one previously unfinished node is finished. This property depends on our particular choice of heuristics.
- As a result, HSVI($\epsilon$) is guaranteed to terminate after performing at most $u_{\max}$ updates, where
$$u_{\max} = t_{\max} \frac{(|\mathcal{A}||\mathcal{O}|)^{t_{\max}+1} - 1}{|\mathcal{A}||\mathcal{O}| - 1}.$$

(Note this is a conservative theoretical bound; empirically, it is much faster.)

## 5 THE *ROCKSAMPLE* PROBLEM

To test HSVI, we have developed *RockSample*, a scalable problem that models rover science exploration (fig. 4). The rover can achieve reward by sampling rocks in the immediate area, and by continuing its traverse (reaching the exit at the right side of the map). The positions of the rover and the rocks are known, but only some of the rocks have scientific value; we will call these rocks "good". Sampling a rock is expensive, so the rover is equipped with a noisy long-range sensor that it can use to help determine whether a rock is good before choosing whether to approach and sample it.

An instance of *RockSample* with map size $n \times n$ and $k$ rocks is described as *RockSample*[$n, k$]. The POMDP model of *RockSample*[$n, k$] is as follows. The state space is the cross product of $k + 1$ features: *Position* $= \{(1,1), (1,2), \ldots, (n,n)\}$, and $k$ binary features *RockType*$_i$ = {*Good*, *Bad*} that indicate which of the rocks are good. There is an additional terminal state, reached when the rover moves off the right-hand edge of the map. The rover can select from $k + 5$ actions: {*North*, *South*, *East*, *West*, *Sample*, *Check*$_1$, ..., *Check*$_k$}. The first four are deterministic single-step motion actions. The *Sample* action samples the rock at the rover's current location. If the rock is good, the rover receives a reward of 10 and the rock becomes bad (indicating that nothing more can be gained by sampling it). If the rock is bad, it receives a penalty of $-10$. Moving into the exit area yields reward 10. All other moves have no cost or reward.

Each *Check*$_i$ action applies the rover's long-range sensor to rock $i$, returning a noisy observation from {*Good*, *Bad*}. The noise in the long-range sensor reading is determined by the efficiency $\eta$, which decreases exponentially as a function of Euclidean distance from the target. At $\eta = 1$, the sensor always returns the correct value. At $\eta = 0$, it has a 50/50 chance of returning *Good* or *Bad*. At intermediate values, these behaviors are combined linearly. The initial belief is that every rock has equal probability of being *Good* or *Bad*.

## 6 EXPERIMENTAL RESULTS

We tested HSVI on several well-known problems from the scalable POMDP literature, as well as instances of *RockSample*. Our benchmark set follows [Pineau et al., 2003], which provides performance numbers for PBVI and some other value iteration algorithms. Note that all of the problems have $\gamma = 0.95$.

Experiments were conducted as follows. Periodically during each run, we interrupted HSVI and simulated its current best policy $\pi$, providing an estimate of the solution quality, $J^\pi(b_0)$. The reported quality is the average reward received over many simulation runs (100-1000). Replicating earlier experiments, each simulation was terminated after 251 steps.

For each problem, results are reported over a single run of the algorithm. In a few cases we made multiple runs, but since HSVI is not stochastic, successive runs are identical up to minuscule changes arising from varying background load on the platform we used, a Pentium-III running at 850 MHz, with 256 MB of RAM.

Fig. 5 shows HSVI solution quality vs. time for four problems. In these plots, we also track the bounds $\underline{V}(b_0)$ and $\bar{V}(b_0)$. Recall that at every phase of the algorithm, we are guaranteed that $\underline{V}(b_0) \leq J^\pi(b_0) \leq \bar{V}(b_0)$. Fig. 5 should reflect this, at least up to the error in our estimate of $J^\pi(b_0)$ (errorbars are 95% confidence intervals). *ThreeState* is a trivial problem we generated, an example of HSVI running

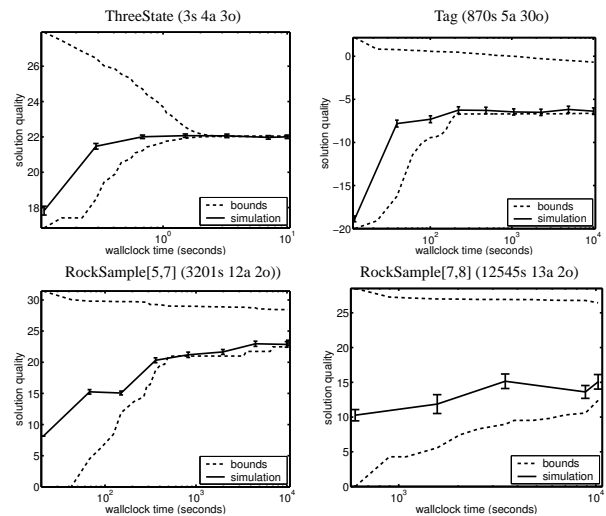

Figure 5: Solution quality vs. time.

to convergence. On the larger problems, the bounds have not converged by the end of the run.

Fig. 6 shows running times and final solution quality for HSVI and some other state-of-the-art algorithms. Unfortunately, not all competitive algorithms could be included in the comparison, because there is no widely accepted POMDP benchmark that we could use. Results for algorithms other than HSVI were computed on different platforms; running times are only very roughly comparable. Among the algorithms compared, HSVI's final solution quality is in every case within measurement error of the best reported so far, and in one case (the *Tag* problem) is significantly better.

| Problem (num. states/actions/observations) | Goal% | Reward | Time (s) | $|\Gamma|$ |
|---|---|---|---|---|
| **Tiger-Grid** (36s 5a 17o) | | | | |
| QMDP [Pineau et al., 2003] | n.a. | 0.198 | 0.19 | n.a. |
| Grid [Brafman, 1997] | n.a. | 0.94 | n.v. | 174 |
| PBUA [Poon, 2001] | n.a. | 2.30 | 12116 | 660 |
| PBVI [Pineau et al., 2003] | n.a. | 2.25 | 3448 | 470 |
| HSVI | n.a. | 2.35 | 10341 | 4860 |
| **Hallway** (61s 5a 21o) | | | | |
| QMDP [Littman et al., 1995] | 47.4 | n.v. | n.v. | n.a. |
| PBUA [Poon, 2001] | 100 | 0.53 | 450 | 300 |
| PBVI [Pineau et al., 2003] | 96 | 0.53 | 288 | 86 |
| HSVI | 100 | 0.52 | 10836 | 1341 |
| **Hallway2** (93s 5a 17o) | | | | |
| QMDP [Littman et al., 1995] | 25.9 | n.v. | n.v. | n.a. |
| Grid [Brafman, 1997] | 98 | n.v. | n.v. | 337 |
| PBUA [Poon, 2001] | 100 | 0.35 | 27898 | 1840 |
| PBVI [Pineau et al., 2003] | 98 | 0.34 | 360 | 95 |
| HSVI | 100 | 0.35 | 10010 | 1571 |
| **Tag** (870s 5a 30o) | | | | |
| QMDP [Pineau et al., 2003] | 17 | -16.769 | 13.55 | n.a. |
| PBVI [Pineau et al., 2003] | 59 | -9.180 | 180880 | 1334 |
| HSVI | 100 | -6.37 | 10113 | 1657 |
| **RockSample[4,4]** (257s 9a 2o) | | | | |
| PBVI [Pineau, personal communication] | n.a. | 17.1 | ~2000 | n.v. |
| HSVI | n.a. | 18.0 | 577 | 458 |
| **RockSample[5,5]** (801s 10a 2o) | | | | |
| HSVI | n.a. | 19.0 | 10208 | 699 |
| **RockSample[5,7]** (3201s 12a 2o) | | | | |
| HSVI | n.a. | 23.1 | 10263 | 287 |
| **RockSample[7,8]** (12545s 13a 2o) | | | | |
| HSVI | n.a. | 15.1 | 10266 | 94 |
| n.a. = not applicable | n.v. = not available | | | |

Figure 6: Multi-algorithm performance comparison.

| Problem (num. states/actions/observations) | $v_c$ | Time PBVI | Time HSVI | Speedup |
|---|---|---|---|---|
| **Tiger-Grid** (36 s 5a 17o) | 2.25 | 3448 | 1053 | 3.3 |
| **Hallway** (61s 5a 21o) | 0.52 | 100-200 | 163 | $\sim$1 |
| **Hallway2** (93s 5a 17o) | 0.34 | 360 | 181 | 2.0 |
| **Tag** (870s 5a 30o) | -9.18 | 180880 | 39 | 4600 |
| **RockSample[4,4]** (256s 9a 2o) | 17.1 | $\sim$2000 | 23 | $\sim$87 |

Figure 7: Performance comparison, HSVI and PBVI.

Fig. 6, which shows only a single time/quality data point for each problem, does not provide enough data for speed comparisons. Therefore we decided to make a closer comparison with one algorithm. PBVI was chosen both because it is a competitive algorithm, and because [Pineau et al., 2003] presents detailed solution quality vs. time curves for our benchmark problems.

In order to control for differing lengths of runs, we report the time that each algorithm took to reach a common value $v_c$, defined to be the highest value that *both* algorithms were able to reach at some point during their run. There is uncertainty associated with some of the times for PBVI because they were derived from manual reading of published plots; this uncertainty is noted in our comparison table. PBVI and HSVI appear to have been run on comparable platforms.[3]

Fig. 7 compares PBVI and HSVI performance. The two algorithms show similar performance on smaller problems. As the problems scale up, however, HSVI provides dramatic speedup. A brief explanation of why this might be the case: Recall that the policy returned by HSVI is based solely on the lower bound. The upper bound is used only to guide forward exploration. But upper bound updates, which involve the solution of several linear programs, often take longer than lower bound updates. Since PBVI keeps only a lower bound, its updates proceed much more quickly. HSVI can only have competitive performance to the extent that the intelligence of its heuristics outweighs the speed penalty of updating the upper bound. Apparently, the heuristics become relatively more important as problem size increases.

## 7 RELATED WORK

Because HSVI combines several existing solution techniques, it can be compared to a wide range of related work. Figure 8, although far from exhaustive, lists many relevant algorithms and some of the features they share with HSVI.

[Hauskrecht, 1997], perhaps the closest prior work, describes separate algorithms for incrementally calculating the upper bound (ICUB) and lower bound (ICLB). The ICUB upper bound is similar to that of HSVI in that it is initialized with the value function for the underlying MDP ($V_{\text{MDP}}$), improved with asynchronous backups, and used as an action heuristic. Unlike HSVI, ICUB uses a grid-based representation, and explores forward from belief space critical points rather than a specified initial belief. The ICUL lower bound uses the same vector set representation as HSVI and adds the result of each gradient backup in the same way. But because ICUB and ICUL are separate algorithms, ICUL's forward exploration does not select actions based on the upper bound, and neither algorithm makes use of an uncertainty-based observation heuristic.

---

[3]PBVI performance on *RockSample*[4, 4] and a rough performance estimate for the computer used in PBVI experiments were provided courtesy of J. Pineau (personal communication).

|  | Applied to POMDPs | Examines only reachable beliefs | Asynchronous updates | Uses observation/outcome heuristic | Uses action heuristic | Leverages value function convexity | Keeps upper and lower bounds |
|---|---|---|---|---|---|---|---|
| HSVI | Y | Y | Y | Y | Y | Y | Y |
| ICUB/ICUL [Hauskrecht, 1997] | Y | Y | - | Y | - | Y | Y |
| BI-POMDP [Washington, 1997] | Y | Y | Y | Y | Y | - | Y |
| RTDP-BEL [Geffner and Bonet, 1998] | Y | Y | Y | Y | - | - | - |
| [Brafman, 1997] | Y | Y | - | Y | Y | Y | - |
| [Dearden and Boutilier, 1994] | - | Y | Y | Y | - | - | Y |
| LAO* [Hansen and Zilberstein, 2001] | - | Y | Y | Y | - | - | Y |
| PBVI [Pineau et al., 2003] | Y | Y | - | Y | - | Y | - |
| PBDP [Zhang and Zhang, 2001] | Y | - | - | - | - | Y | - |
| Incremental pruning [Cassandra et al., 1997] | Y | - | - | - | - | Y | - |

Figure 8: Relevant algorithms and features.

Other related work mostly falls into two camps. The first are algorithms that combine heuristic search with dynamic programming updates. RTDP-BEL [Geffner and Bonet, 1998], a POMDP extension of the well-known RTDP value iteration technique for MDPs [Barto et al., 1995], turns out to be very similar to ICUB. BI-POMDP [Washington, 1997] uses forward exploration based on AO* with $V_{\text{MDP}}$ as its heuristic. BI-POMDP keeps upper and lower bounds on nodes in the search tree—however, it does not explicitly represent the bounds as functions, so it is unable to generalize the value at a belief to neighboring beliefs. Some other algorithms in this group are [Dearden and Boutilier, 1994, Brafman, 1997, Hansen and Zilberstein, 2001].

The second camp includes algorithms that employ a piecewise linear convex value function representation and gradient backups. There are a host of algorithms along these lines, dating back to [Sondik, 1971]. Most differ from HSVI in that they perform gradient backups over the full belief space instead of focusing on relevant beliefs. One exception is PBVI [Pineau et al., 2003], which performs synchronous gradient backups on a growing subset of the belief space, designed such that it examines only reachable beliefs. Unlike HSVI, PBVI does not keep an upper bound and does not use a value-based action heuristic when expanding its belief set. Other algorithms in this group include incremental pruning [Cassandra et al., 1997] and point-based dynamic programming [Zhang and Zhang, 2001].

HSVI avoids examining unreachable beliefs using forward exploration. [Boutilier et al., 1998] describe how to precompute reachability in order to eliminate states in an MDP context. In a POMDP context their technique would go beyond HSVI by explicitly reducing the dimensionality of the belief space, but the remaining space might still include unreachable beliefs never visited by HSVI.

Finally, there are many competitive POMDP solution approaches that do not employ heuristic search or a PWLC value function representation: too many to discuss here. We refer the reader to a survey [Aberdeen, 2002]. Hopefully, increased adoption of common benchmarks in the POMDP community will allow us to better compare HSVI with other algorithms in the future.

## 8   CONCLUSION

This paper presents HSVI, a POMDP solution algorithm that uses heuristics, based on upper and lower bounds of the optimal value function, to guide local updates.

Experimentally, HSVI is able to find solutions with quality within measurement error of the best previous report on all of the benchmark problems we tried, and it did significantly better on the *Tag* problem. In time comparisons with the state-of-the-art PBVI algorithm, HSVI showed dramatic speedups on larger problems. We applied HSVI to an instance of the new *RockSample* domain with 12,545 states, more than 10 times larger than most problems presented in the scalable POMDP literature.

There are several ways that we would like to extend HSVI. First, it should be possible to speed up lower bound updates through the following observation: most beliefs are sparse, and most $\alpha$ vectors are optimal for only a few closely related beliefs. Therefore, only a few elements of any given $\alpha$ vector are relevant, and we can save effort if we avoid computing the rest. Second, we are working on reducing the number of LP calculations needed for the upper-bound by pruning some actions early, and by reusing old LP solutions. Finally, we could leverage better data structures such as ADDs for representing beliefs, $\alpha$ vectors, and other objects used by the algorithm [Hoey et al., 1999].

In summary, this is an exciting time: recent progress in solution performance suggests that the POMDP planning model will soon be a feasible choice for robot decision-making on a much wider range of real problems.